\documentclass[sigconf]{acmart}

\usepackage{algorithm}
\usepackage{algorithmic}
\usepackage{multirow}
\usepackage{enumitem}
\usepackage{caption}
\usepackage{booktabs}
\usepackage{amsmath,amsfonts}
\usepackage{textcomp}
\usepackage{subcaption}
\usepackage{footnote}
\usepackage{graphicx}

\allowdisplaybreaks[4]
\usepackage{hyperref}

\theoremstyle{definition}

\AtBeginDocument{%
  \providecommand\BibTeX{{%
    \normalfont B\kern-0.5em{\scshape i\kern-0.25em b}\kern-0.8em\TeX}}}

\copyrightyear{2026}
\acmYear{2026}
\setcopyright{cc}
\setcctype{by}
\acmConference[KDD '26]{Proceedings of the 32nd ACM SIGKDD Conference on Knowledge Discovery and Data Mining V.2}{August 09--13, 2026}{Jeju Island, Republic of Korea}
\acmBooktitle{Proceedings of the 32nd ACM SIGKDD Conference on Knowledge Discovery and Data Mining V.2 (KDD '26), August 09--13, 2026, Jeju Island, Republic of Korea}
\acmDOI{10.1145/3770855.3818435}
\acmISBN{979-8-4007-2259-2/2026/08}
\begin{document}

\title{HOBA: Hierarchical On-Policy Bidding Agents for Adaptive Online Advertising}

\author{Ji Wu}
\affiliation{
  \institution{Kuaishou Technology}
  \city{Beijing}
  \country{China}
}
\email{wuji03@kuaishou.com}

\author{Yunshan Peng}
\affiliation{
  \institution{Kuaishou Technology}
  \city{Beijing}
  \country{China}
}
\email{pengyunshan@kuaishou.com}

\author{Wentao Bai}
\affiliation{
  \institution{Kuaishou Technology}
  \city{Beijing}
  \country{China}
}
\email{baiwentao@kuaishou.com}

\author{Yunke Bai}
\affiliation{
  \institution{Kuaishou Technology}
  \city{Beijing}
  \country{China}
}
\email{baiyunke@kuaishou.com}

\author{Wenzheng Shu}
\affiliation{
  \institution{Kuaishou Technology}
  \city{Beijing}
  \country{China}
}
\email{shuwenzheng@kuaishou.com}

\author{Jinan Pang}
\authornote{J.Pang is the corresponding author.}
\affiliation{
  \institution{Kuaishou Technology}
  \city{Beijing}
  \country{China}
}
\email{pangjinan@kuaishou.com}
\author{Yanxiang Zeng}
\affiliation{
  \institution{Kuaishou Technology}
  \city{Beijing}
  \country{China}
}
\email{zengyanxiang@kuaishou.com}

\author{Xialong Liu}
\affiliation{
  \institution{Kuaishou Technology}
  \city{Beijing}
  \country{China}
}
\email{zhaolei16@kuaishou.com}

\renewcommand{\shortauthors}{Ji Wu et al.}

\begin{abstract}
Online advertising bidding systems typically deploy multiple offline-trained 
expert models (e.g., PID controllers, model predictive control, offline RL 
policies) but face two critical limitations: lack of online adaptability to 
non-stationary auction markets, and reliance on costly manual tuning of 
hyperparameters such as bid bounds and budget pacing constraints.
We propose HOBA (Hierarchical On-policy Bidding Agents), a hierarchical 
reinforcement learning framework that decouples strategic reasoning, model 
selection, and bid execution across three time scales. At the high level, 
a large language model infers hyperparameters from contextual signals through 
a Think-Act-Observe-Reflect loop with historical experience retrieval. At the 
mid level, a SARSA agent dynamically selects among expert models, incorporating 
causal adjustment to eliminate selection bias. At the low level, a dynamic 
expert pool (PID, MPC, IQL, Decision Transformer) executes bids under high-level 
constraints. This design confines online learning to discrete expert selection 
rather than continuous bid optimization, significantly reducing exploration 
risk while maintaining adaptability.
Experiments on the AuctionNet benchmark and a large-scale A/B test demonstrate 
consistent improvements over state-of-the-art baselines. In a large-scale online 
deployment, HOBA delivered substantial business value, achieving a +3.6\% increase 
in target cost, proving the effectiveness of our hierarchical multi-agent bidding 
paradigm.
\end{abstract}

\ccsdesc[500]{Information systems~Online advertising}
\keywords{Online Advertising, Auto-bidding, Reinforcement Learning, Agent, LLM}

\maketitle
\section{Introduction}

Online advertising auctions constitute a multi-billion dollar market where advertisers compete in real-time to maximize objectives (conversions, revenue) subject to budget and cost-per-acquisition (CPA) constraints. A core challenge is \textbf{automatic bidding} in non-stationary markets where competitor strategies, user behavior, and platform dynamics continuously evolve~\cite{hu2022applying}, necessitating systems that adapt online while maintaining safety.

Current industry practice deploys offline-trained expert models—PID controllers~\cite{weinan2018bid}, model predictive control (MPC)~\cite{han2020reinforcement}, and offline RL policies~\cite{kumar2020conservative,kostrikov2021offline,chen2021decision}—as bidding controllers. However, these systems face two critical limitations: (1) \textbf{lack of online adaptability}, as expert parameters remain fixed post-deployment and cannot respond to market shifts without costly retraining, and (2) \textbf{manual hyperparameter tuning} of bid bounds, pacing rates, and exploration coefficients per campaign. Enabling safe online learning remains challenging: on-policy methods (PPO~\cite{schulman2017proximal}, A3C~\cite{mnih2016asynchronous}) risk catastrophic exploration that can exhaust budgets within minutes, while off-policy methods suffer from action collapse and fail to generalize across diverse conditions. Moreover, existing approaches conflate strategic planning with tactical execution, treating bidding as a monolithic optimization problem that lacks interpretability.

We propose \textbf{HOBA (Hierarchical On-policy Bidding Agents)}, a framework that decomposes bidding into three tiers: (1) a \textbf{high-level LLM agent} (1-hour cycle) infers hyperparameters $\boldsymbol{\theta}_t$ through Think-Act-Observe-Reflect with experience retrieval, (2) a \textbf{mid-level SARSA agent} (2-minute cycle) selects experts using causal adjustment to debias historical data, and (3) a \textbf{low-level expert pool} (per-auction) executes bids under high-level constraints. We treat each layer as an autonomous agent with its own observation space, decision cycle, and optimization objective, forming a cooperative multi-agent system under a shared revenue constraint.
This design confines online learning to \textbf{discrete expert selection} (5--10 validated models) rather than continuous bid optimization, achieving safe adaptation without unconstrained exploration.

Our contributions are: \textbf{(1)} We propose the first hierarchical multi-agent bidding framework decoupling strategic reasoning, expert selection, and bid execution across time scales. \textbf{(2)} We introduce LLM-guided hyperparameter optimization with natural language reasoning and causal-adjusted on-policy expert selection for safe online adaptation. \textbf{(3)} We demonstrate +3.6\% cost target improvement  in large-scale A/B tests, validating practical viability.

\section{Related Work}
\subsection{Automatic Bidding in Online Advertising}

Automatic bidding systems aim to optimize advertiser objectives under budget and cost constraints. Early approaches employed rule-based controllers: \textbf{PID controllers}~\cite{weinan2018bid} adjust bid multipliers via proportional-integral-derivative feedback to regulate CPA deviation, while \textbf{model predictive control (MPC)}~\cite{han2020reinforcement} solves constrained optimization over future horizons to balance conversions and budget consumption. These methods offer interpretability and fast deployment but require extensive manual tuning and struggle to adapt to non-stationary markets.

Recent work has increasingly adopted \textbf{reinforcement learning} for bidding. \citet{cai2017real} formulate bidding as an MDP and apply Q-learning to optimize real-time display advertising. \citet{han2020reinforcement} introduce a generalized framework optimizing diverse KPIs via constrained policy gradient methods. However, these on-policy approaches face safety risks in production deployment—unconstrained exploration can exhaust budgets or violate cost targets within minutes, causing advertiser churn.

To enable safe learning, practitioners have turned to \textbf{offline RL}~\cite{levine2020offline}, training policies on logged data without online exploration. Recent work has applied Conservative Q-Learning (CQL)~\cite{kumar2020conservative}  to mitigate overestimation in bidding, employed Implicit Q-Learning (IQL)~\cite{kostrikov2021offline} for robust value estimation under reward noise, and leveraged Decision Transformer (DT)~\cite{chen2021decision} to model sequential dependencies via supervised sequence modeling. While these methods avoid risky exploration, they suffer from action collapse—policies fail to generalize across diverse market conditions due to distributional shift between offline data and online deployment~\cite{fu2021benchmarks}.

Our work differs fundamentally by \textbf{decoupling expert training from online adaptation}: we maintain a pool of diverse offline-trained experts and confine online learning to discrete expert selection rather than continuous bid optimization, achieving safety without sacrificing adaptability.

\subsection{Hierarchical Reinforcement Learning}

Hierarchical RL decomposes complex tasks into multiple levels of abstraction, enabling efficient learning and transfer~\cite{barto2003recent,pateria2021hierarchical}. \textbf{Options framework}~\cite{sutton1999between} extends MDPs with temporally extended actions, while \textbf{feudal RL}~\cite{dayan1992feudal,vezhnevets2017feudal} separates managers (setting subgoals) from workers (achieving subgoals). \citet{nachum2018data} propose hierarchical off-policy learning with goal-conditioned value functions. In robotics, \citet{osa2020algorithmic} survey hierarchical methods for manipulation tasks requiring coordination across motion primitives.

In online advertising, prior work has applied hierarchical methods mainly to budget allocation across campaigns or coordination among multiple advertisers~\cite{yuan2021hierarchical}, treating individual bidders as monolithic policies. In contrast, HOBA decomposes a \emph{single} advertiser's bidding system into three hierarchical levels: strategic reasoning (LLM for hyperparameter inference), tactical selection (SARSA for expert model choice), and operational execution (expert pool for bid generation). This time-scale separation confines online learning to discrete model selection rather than continuous bid optimization, significantly reducing exploration risk.

\subsection{LLM-Based Decision Making and Agent Systems}
Large language models have emerged as powerful tools for sequential decision-making through natural language reasoning~\cite{huang2022language,yao2023react}. Methods like ReAct~\cite{yao2023react} combine reasoning with action execution, while Reflexion~\cite{shinn2023reflexion} enables self-improvement through experience replay. Recent work has applied LLMs to hyperparameter tuning~\cite{yang2024large} and AutoML configuration~\cite{chen2023llm}, demonstrating superior sample efficiency in offline settings. However, these approaches focus on simulated environments or offline optimization. HOBA is the first to deploy LLM-guided reasoning for \textbf{online hyperparameter inference} in a production advertising system, where a Think-Act-Observe-Reflect loop with historical experience retrieval enables continual adaptation to non-stationary markets without per-auction LLM inference costs.

\subsection{Causal Inference in Reinforcement Learning}
Off-policy learning from logged data faces confounding bias when certain actions are preferentially chosen in favorable contexts~\cite{li2015toward,swaminathan2015batch}. Inverse propensity scoring (IPS)~\cite{horvitz1952generalization} reweights samples by logging policy probabilities but suffers from high variance, while doubly robust estimation~\cite{bang2005doubly,dudik2014doubly} combines model-based prediction with importance weighting for improved robustness. In advertising, prior work has applied these methods to click-through rate prediction~\cite{li2010contextual} and ranking evaluation~\cite{gilotte2018offline}. Our mid-level SARSA agent incorporates doubly robust causal adjustment (Eq.~\ref{eq:doubly_robust}) to debias expert selection, ensuring Q-learning does not inherit spurious correlations from historical data where expert models were deployed under non-random policies.

\section{Preliminary}
We formulate automatic bidding as a sequential decision problem where an advertiser aims to maximize conversion value subject to budget and cost constraints over a campaign horizon.

\subsection{Auto-Bidding as Constrained Optimization}

For advertiser $i$ participating in auctions $j \in \{1, \ldots, J\}$, the bidding problem is formulated as~\cite{han2020reinforcement}:
\begin{equation}
\label{eq:bidding_objective}
\begin{aligned}
\max_{b_i} \sum_{j} \mathcal{A}_{ij}(b_j) \cdot v_{ij} \quad \text{s.t.} \quad & \sum_{j} \mathcal{A}_{ij}(b_j) \cdot \mathcal{P}_{ij} \leq B_i, \\
& \frac{\sum_{j} \mathcal{A}_{ij}(b_j) \cdot \mathcal{P}_{ij}}{\sum_{j} \mathcal{A}_{ij}(b_j) \cdot v_{ij}} \leq C_i,
\end{aligned}
\end{equation}
where $b_j$ denotes bids from all advertisers, $v_{ij} \in \mathbb{R}_+$ is conversion value, $\mathcal{A}_{ij}(b_j) \in \{0, 1\}$ indicates whether advertiser $i$ wins auction $j$, and $\mathcal{P}_{ij}$ is payment upon winning. The constraints enforce a budget limit $B_i$ and maximum cost-per-value ratio $C_i$ (equivalently, minimum return-on-ad-spend).

\subsection{Challenges in Direct Policy Learning}
Standard RL approaches face fundamental difficulties in online advertising:

\textbf{Safety-Adaptability Tradeoff.} On-policy methods (e.g., PPO~\cite{schulman2017proximal}, A3C~\cite{mnih2016asynchronous}) require online exploration but risk catastrophic failures—a single poorly-informed bid can exhaust the daily budget within minutes. Off-policy methods (e.g., CQL~\cite{kumar2020conservative}, IQL~\cite{kostrikov2021offline}) train safely on logged data but suffer from distributional shift: policies fail when market conditions (competitor behavior, traffic patterns) deviate from historical data~\cite{levine2020offline}.

\textbf{Hyperparameter Sensitivity.} Performance depends critically on hyperparameters such as bid bounds, budget pacing rates, and exploration bonuses. Manual tuning is labor-intensive and campaign-specific, while automated methods like Bayesian optimization~\cite{snoek2012practical} require hundreds of evaluations—infeasible when each evaluation involves deploying a policy for an entire campaign day.

\textbf{Interpretability Requirements.} Production systems demand transparency: advertisers need to understand bidding decisions, and platform operators must audit constraint compliance. End-to-end neural policies provide limited interpretability, complicating debugging and trust.

\subsection{Hierarchical Decomposition Rationale}
HOBA addresses these challenges through hierarchical decomposition across three time scales:
\begin{itemize}[leftmargin=*,noitemsep]
\item \textbf{Strategic Layer (hourly)}: An LLM infers hyperparameters from campaign context via natural language reasoning, separating \textit{what constraints to impose} from \textit{how to bid}.
\item \textbf{Tactical Layer (minute-scale)}: A SARSA agent selects among pre-validated expert models. By learning over discrete expert choices ($|\mathcal{A}| \approx 5$-$10$) rather than continuous bid values, exploration risk is dramatically reduced.
\item \textbf{Operational Layer (per-auction)}: Offline-trained experts execute bids under high-level constraints, enabling fast inference without online policy updates.
\end{itemize}
This design achieves: (1) \textbf{Safety} via bounded exploration, (2) \textbf{Adaptability} via online expert selection responding to market shifts, and (3) \textbf{Interpretability} via LLM reasoning traces and causal attribution. Formal details follow in Section~\ref{sec:method}.

\section{Methodology}
\label{sec:method}
\begin{figure}[t]
\centering
\includegraphics[width=0.48\textwidth]{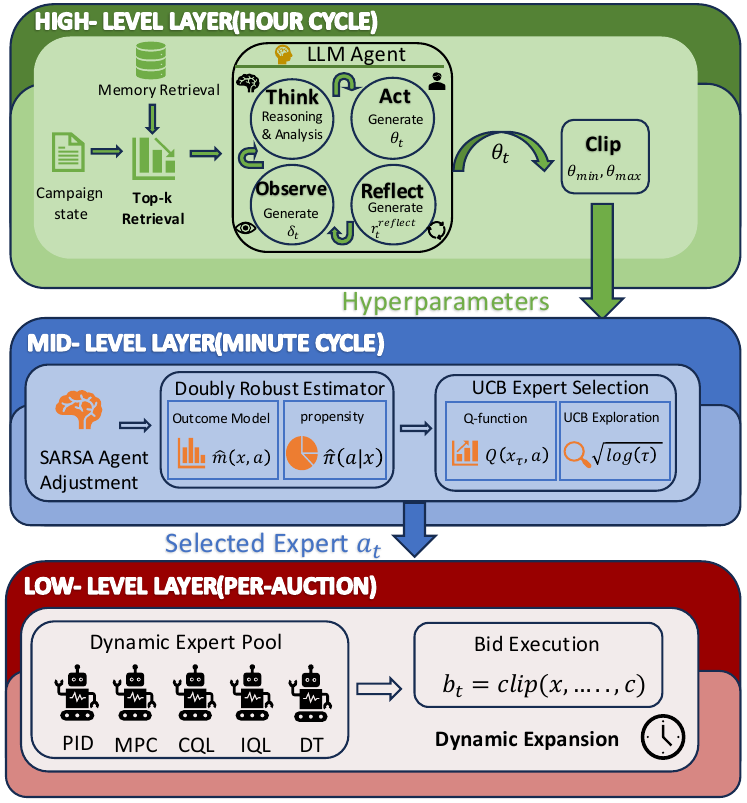}
\caption{Overview of HOBA's three-tier architecture. High-level LLM agent (hour cycle) generates hyperparameters $\boldsymbol{\theta}_t$ via Think-Act-Observe-Reflect; Mid-level SARSA agent (minute cycle) selects expert models $a_\tau$ with causal adjustment; Low-level expert pool (per-auction) executes bids $b_\tau$ with fixed policies. Online learning is confined to mid-level selection, ensuring safe adaptation.}
\label{fig:framework}
\end{figure}

HOBA consists of three hierarchical tiers operating at different time scales (Figure~\ref{fig:framework}): (1) a high-level LLM agent making strategic decisions hourly, (2) a mid-level SARSA agent selecting expert models every 2 minutes, and (3) a low-level expert pool executing bids per auction.This decomposition confines online learning to discrete expert selection rather than continuous bid optimization, reducing exploration risk while maintaining adaptability.
\noindent\textbf{Global Objective Alignment.}
All three layers optimize the single global objective in 
Eq.~(1). The LLM parameterizes the feasible constraint set; 
SARSA selects the best expert within that region; and experts 
enforce constraints by construction via double-clipping 
(Eq.~(15)), guaranteeing every bid satisfies Eq.~(1). 
The penalties $\mu$ and $\eta$ in Eq.~(11) are instrumental: 
switching instability causes budget waste (violating $B_i$), 
and pacing deviation causes CPA violations (violating $C_i$).

\subsection{High-Level: LLM-Guided Hyperparameter Optimization}
\label{subsec:high_level}

The high-level agent indirectly shapes the MDP of the mid-level agent by dynamically constraining the feasible action space and reward scale.The LLM agent infers optimal hyperparameters from campaign context through a Think-Act-Observe-Reflect loop~\cite{yao2023react,shinn2023reflexion}, operating on a 1-hour cycle to reduce exploration risk while enabling strategic adaptation.

\subsubsection{State Representation and Decision Loop}
At time $t$, we construct high-level state capturing budget status ($\mathbf{b}_t$), performance metrics ($\mathbf{r}_t$), market conditions ($\mathbf{m}_t$), placement performance ($\mathbf{p}_t$), and historical statistics ($\mathbf{h}_t$):
\begin{equation}
\label{eq:state_high}
\mathbf{s}_t^H = [\mathbf{b}_t; \mathbf{r}_t; \mathbf{m}_t; \mathbf{p}_t; \mathbf{h}_t].
\end{equation}
This numeric vector is transformed into natural language via template $\mathcal{P}_{\text{state}}(\mathbf{s}_t^H)$ for LLM processing.

\textbf{Think Phase.} The LLM retrieves similar historical experiences from memory bank $\mathcal{D}_{\text{mem}}$ and performs structured reasoning:
\begin{equation}
\label{eq:think}
\mathbf{a}^{\text{think}}_t = \text{LLM}(\mathcal{P}_{\text{think}}(\mathbf{s}_t^H, \mathcal{M}_t)),
\end{equation}
where $\mathcal{M}_t$ contains top-$k$ retrieved experiences based on semantic similarity, producing problem identification and root cause analysis.

\textbf{Act Phase.} Based on analysis, the LLM generates hyperparameters:
\begin{equation}
\label{eq:act}
\boldsymbol{\theta}_t = \text{LLM}(\mathcal{P}_{\text{act}}(\mathbf{s}_t^H, \mathbf{a}^{\text{think}}_t)),
\end{equation}
where $\boldsymbol{\theta}_t = \{\theta^{\text{bounds}}_t, \theta^{\text{mode}}_t, \theta^{\text{control}}_t, \theta^{\text{explore}}_t\}$ includes bid bounds $[q_{\text{lower}}, q_{\text{upper}}]$, response mode (conservative/moderate/aggressive), control targets (CPA, budget pacing rate $\rho$, bid change limits $\boldsymbol{\Delta}$), and exploration bonus. Safety is enforced via:
\begin{equation}
\label{eq:clip}
\boldsymbol{\theta}_t \leftarrow \text{Clip}(\boldsymbol{\theta}_t, \boldsymbol{\theta}_{\min}, \boldsymbol{\theta}_{\max}).
\end{equation}

\textbf{Observe-Reflect.} After one-hour execution, we aggregate observations and compute deviation from predictions:
\begin{equation}
\label{eq:observe}
\mathbf{o}_t = \text{Aggregate}(\{\mathbf{x}_{\tau}\}_{\tau \in \mathcal{T}_t}), \quad \boldsymbol{\delta}_t = \{\delta_{\text{CPA}}, \delta_{\text{conv}}, \delta_{\text{budget}}\}.
\end{equation}
The LLM evaluates decision quality:
\begin{equation}
\label{eq:reflect}
\mathbf{r}^{\text{reflect}}_t = \text{LLM}(\mathcal{P}_{\text{reflect}}(\mathbf{a}^{\text{think}}_t, \boldsymbol{\theta}_t, \mathbf{o}_t, \boldsymbol{\delta}_t)),
\end{equation}
producing prediction accuracy $s_{\text{acc}}$, overall quality score:
\begin{equation}
\label{eq:quality}
s_{\text{quality}} = w_1 s_{\text{objective}} + w_2 s_{\text{acc}} + w_3 s_{\text{risk}} + w_4 s_{\text{efficiency}},
\end{equation}
and structured lessons. This experience is stored in $\mathcal{D}_{\text{mem}}$ as:
\begin{equation}
\label{eq:memory_entry}
e_i = (\mathbf{s}_i^H, \boldsymbol{\theta}_i, \mathbf{o}_i, \mathbf{r}^{\text{reflect}}_i, \mathbf{z}_i),
\end{equation}
with semantic embedding $\mathbf{z}_i$ via pre-trained encoders~\cite{reimers2019sentence}. Retrieval uses cosine similarity with FAISS indexing~\cite{johnson2019billion}:
\begin{equation}
\label{eq:memory_retrieval}
\mathcal{M}_t = \text{TopK}\left(\left\{\frac{\mathbf{z}_t^\top \mathbf{z}_i}{\|\mathbf{z}_t\| \|\mathbf{z}_i\|}\right\}_{i=1}^{|\mathcal{D}_{\text{mem}}|}, k\right).
\end{equation}
Algorithm~\ref{algo:high_level} summarizes the procedure.

\begin{algorithm}[t]
\caption{High-Level LLM Agent in HOBA}
\label{algo:high_level}
\begin{algorithmic}[1]
\STATE \textbf{Input:} Campaign state $\mathbf{s}_t^H$, memory bank $\mathcal{D}_{\text{mem}}$
\STATE \textbf{Output:} Hyperparameters $\boldsymbol{\theta}_t$, reflection $\mathbf{r}_t^{\text{reflect}}$
\STATE 
\STATE // Think Phase
\STATE $\mathbf{z}_t \leftarrow \text{Embed}(\text{StateToText}(\mathbf{s}_t^H))$
\STATE $\mathcal{M}_t \leftarrow \text{TopK}(\{\text{sim}(\mathbf{z}_t, \mathbf{z}_i)\}_{i \in \mathcal{D}_{\text{mem}}}, k=3)$
\STATE $\mathbf{a}_t^{\text{think}} \leftarrow \text{LLM}(\mathcal{P}_{\text{think}}(\mathbf{s}_t^H, \mathcal{M}_t))$
\STATE 
\STATE // Act Phase
\STATE $\boldsymbol{\theta}_t, \mathbf{o}_t^{\text{expected}} \leftarrow \text{LLM}(\mathcal{P}_{\text{act}}(\mathbf{s}_t^H, \mathbf{a}_t^{\text{think}}))$
\STATE $\boldsymbol{\theta}_t \leftarrow \text{Clip}(\boldsymbol{\theta}_t, \boldsymbol{\theta}_{\min}, \boldsymbol{\theta}_{\max})$
\STATE 
\STATE // Observe Phase 
\STATE $\mathbf{o}_t \leftarrow \text{Aggregate}(\text{ExecuteOneHour}(\boldsymbol{\theta}_t))$
\STATE $\boldsymbol{\delta}_t \leftarrow \text{ComputeDeviation}(\mathbf{o}_t, \mathbf{o}_t^{\text{expected}})$
\STATE 
\STATE // Reflect Phase
\STATE $\mathbf{r}_t^{\text{reflect}} \leftarrow \text{LLM}(\mathcal{P}_{\text{reflect}}(\mathbf{a}_t^{\text{think}}, \boldsymbol{\theta}_t, \mathbf{o}_t, \boldsymbol{\delta}_t))$
\STATE $\mathbf{z}_t' \leftarrow \text{Embed}(\text{ExperienceToText}(\mathbf{s}_t^H, \boldsymbol{\theta}_t, \mathbf{o}_t, \mathbf{r}_t^{\text{reflect}}))$
\STATE $\mathcal{D}_{\text{mem}} \leftarrow \mathcal{D}_{\text{mem}} \cup \{(\mathbf{s}_t^H, \boldsymbol{\theta}_t, \mathbf{o}_t, \mathbf{r}_t^{\text{reflect}}, \mathbf{z}_t')\}$
\STATE 
\STATE \textbf{return} $\boldsymbol{\theta}_t$, $\mathbf{r}_t^{\text{reflect}}$
\end{algorithmic}
\end{algorithm}

\subsubsection{Prompt Engineering and Output Validation}
Prompt effectiveness is critical for LLM-guided optimization. Our prompts incorporate: (1) role definition establishing domain expertise (``You are an expert advertising optimization agent''), (2) structured input with formatted performance metrics and placement breakdowns, (3) explicit reasoning framework (problem $\to$ cause $\to$ strategy), and (4) JSON schema enforcement ensuring valid outputs with strict type constraints.

To ensure reliability, we validate LLM outputs through multiple checks. First, we verify JSON parsing success and schema compliance (all required fields present with correct types). Second, we apply domain constraints: bid bounds must satisfy $0.1 \leq q_{\text{lower}} < q_{\text{upper}} \leq 3.0$, pacing rate $\rho \in [0.5, 1.5]$, and exploration bonus $\theta^{\text{explore}} \in [0, 0.5]$. Third, we implement fallback mechanisms: if parsing fails or constraints are violated, we retry with error feedback up to 3 times before reverting to the previous valid configuration. This multi-layer validation ensures system robustness even when LLM outputs are occasionally malformed. Critically, 
all hyperparameter outputs are hard-clipped via Eq.~(5), 
ensuring stochasticity in the LLM reasoning trace does not 
propagate to bid execution.

\textbf{Model Optimization.} We optionally fine-tune the LLM offline using Kahneman-Tversky Optimization (KTO)~\cite{ethayarajh2024kto}, labeling configurations as desirable ($s_{\text{quality}} \geq 7$) or undesirable ($s_{\text{quality}} < 4$) based on deployment outcomes. KTO optimizes by pushing desirable configurations above an average reward baseline while pulling undesirable ones below, naturally encoding loss aversion that aligns with constraint-sensitive bidding. Fine-tuning occurs weekly/monthly with gradual A/B rollout to production.

\subsection{Mid-Level: SARSA-Based Expert Selection with Causal Adjustment}
\label{subsec:mid_level}

The mid-level agent selects expert $a_\tau \in \mathcal{A} = \{\text{PID}, \text{MPC}, \text{DT}, \text{IQL}, \text{CQL}\}$ every 2 minutes based on context $\mathbf{x}_\tau = [\mathbf{x}^{\text{time}}_\tau; \mathbf{x}^{\text{market}}_\tau; \mathbf{x}^{\text{perf}}_\tau; \boldsymbol{\theta}_t; \mathbf{x}^{\text{expert}}_\tau]$, encoding temporal features, market conditions, recent performance, high-level hyperparameters, and per-expert statistics. Reward balances value, cost, and stability:
\begin{equation}
\label{eq:reward}
r_\tau = \text{value}_\tau - \lambda \cdot \text{cost}_\tau - \mu \cdot \mathbb{I}[a_\tau \neq a_{\tau-1}] - \eta \cdot |\text{spend\_rate}_\tau - \text{target\_rate}_\tau|,
\end{equation}
with switching penalty $\mu = 0.1$ and pacing penalty $\eta = 0.05$.

\subsubsection{Causal Adjustment and SARSA Selection}
Historical data $\mathcal{D}_{\text{hist}} = \{(\mathbf{x}_i, a_i, r_i)\}$ suffers from confounding: experts selected in favorable contexts appear superior due to preferential selection. We estimate causal effect $\mu(a) = \mathbb{E}[r | do(a)]$ via doubly robust estimator~\cite{bang2005doubly,dudik2014doubly}:
\begin{equation}
\label{eq:doubly_robust}
\hat{\mu}(a) = \frac{1}{N} \sum_{i=1}^N \left[ \hat{m}(\mathbf{x}_i, a) + \frac{\mathbb{I}(a_i = a)}{\hat{\pi}(a | \mathbf{x}_i)} (r_i - \hat{m}(\mathbf{x}_i, a)) \right],
\end{equation}
where outcome model $\hat{m}(\mathbf{x}, a)$ (XGBoost with 500 trees) predicts $\mathbb{E}[r | \mathbf{x}, a]$ and propensity model $\hat{\pi}(a | \mathbf{x})$ (logistic regression with L2 regularization) estimates logging policy probabilities. Both models are trained daily on the last 7 days of data (~500K samples) with clipped propensity scores $\hat{\pi}(a|\mathbf{x}) \geq 0.05$ to prevent extreme importance weights.

We maintain linear Q-function $Q(\mathbf{x}, a; \boldsymbol{\phi}_a) = \mathbf{w}_a^\top \phi(\mathbf{x})$ with polynomial features (degree 2). Expert selection combines Q-values, causal effects, and UCB exploration:
\begin{equation}
\label{eq:expert_selection}
a_\tau = \arg\max_{a \in \mathcal{A}} \left[ Q(\mathbf{x}_\tau, a) + \hat{\mu}(a) + \theta^{\text{explore}}_t \cdot \beta \sqrt{\frac{\log(\tau)}{N_\tau(a) + 1}} \right],
\end{equation}
where $N_\tau(a)$ counts selections, $\beta = 2.0$ controls exploration, and $\theta^{\text{explore}}_t$ from high-level modulates intensity. After observing transition $(\mathbf{x}_\tau, a_\tau, r_\tau, \mathbf{x}_{\tau+1})$ and selecting next action $a_{\tau+1}$, we update via SARSA~\cite{sutton2018reinforcement}:
\begin{equation}
\label{eq:sarsa_update}
\boldsymbol{\phi}_{a_\tau} \leftarrow \boldsymbol{\phi}_{a_\tau} - \alpha \nabla_{\boldsymbol{\phi}_{a_\tau}} \left( Q(\mathbf{x}_\tau, a_\tau) - [r_\tau + \gamma Q(\mathbf{x}_{\tau+1}, a_{\tau+1})] \right)^2,
\end{equation}
with $\alpha = 0.01$ and $\gamma = 0.99$. Critically, this updates only mid-level Q-network weights; low-level expert models remain fixed.

To handle market non-stationarity, we detect distribution shifts via Kolmogorov-Smirnov test~\cite{massey1951kolmogorov} on reward CDFs. When $D_{KS}$ exceeds threshold ($p < 0.05$), we temporarily double exploration and trigger expert fine-tuning.

\subsection{Low-Level: Dynamic Expert Pool}
\label{subsec:low_level}

The pool contains $|\mathcal{A}| \in [5, 10]$ experts executing bids per auction. \textbf{Expert parameters remain fixed during deployment}—online learning is confined to mid-level selection, ensuring safe adaptation.

\subsubsection{Expert Pool Composition}
We initialize with five diverse offline-trained experts: \textbf{PID} (feedback control with gains $K_p = 0.5$, $K_i = 0.1$, $K_d = 0.05$), \textbf{MPC} (constrained optimization over 1-hour horizon via OSQP~\cite{stellato2020osqp}), and three offline RL models—\textbf{CQL}~\cite{kumar2020conservative} (conservative Q-learning with penalty $\alpha = 5.0$), \textbf{IQL}~\cite{kostrikov2021offline} (expectile regression with $\tau = 0.7$), and \textbf{DT}~\cite{chen2021decision} (decision transformer with context length 20). These experts exhibit complementary strengths (Table~\ref{tab:expert_models}): PID for fast error correction, MPC for explicit budget planning, and offline RL for learning from logged data.

\begin{table}[h]
\centering
\small
\begin{tabular}{lll}
\toprule
\textbf{Expert} & \textbf{Core Mechanism} & \textbf{Key Strength} \\
\midrule
PID & Feedback control & Fast error correction \\
MPC & Constrained optimization & Explicit budget planning \\
CQL/IQL & Offline RL (value-based) & Learn from logged data \\
DT & Sequence modeling & Long-term credit assignment \\
\bottomrule
\end{tabular}
\caption{Expert characteristics enabling adaptation across market conditions.}
\label{tab:expert_models}
\end{table}

To adapt to market shifts, we dynamically expand the pool through fine-tuned variants and hybrid ensembles. Every 24 hours, we identify the best performer and create a fine-tuned variant initialized from its parameters and trained for 100 gradient steps on recent data. We also create weighted combinations $b^{\text{hybrid}}_\tau = w \cdot b^{(a_1)}_\tau + (1 - w) \cdot b^{(a_2)}_\tau$ of complementary experts, where ensemble weight $w$ is updated online via gradient ascent on expected reward. Underperforming dynamic experts are pruned when average reward falls below $90\%$ of the best static expert.

\subsubsection{Bid Execution and Constraints}
Given activated expert $a_\tau$ and auction features $\mathbf{f}_\tau$, the final bid enforces hierarchical constraints:
\begin{equation}
\label{eq:bid_execution}
\begin{aligned}
\tilde{b}_\tau &= \text{Expert}_{a_\tau}(\mathbf{f}_\tau, \mathbf{s}_\tau; \boldsymbol{\phi}_{a_\tau}), \\
b_\tau &= \text{Clip}\big(\text{Clip}(\tilde{b}_\tau, q_{\text{lower}} \cdot \text{CPA}_{\text{target}}, q_{\text{upper}} \cdot \text{CPA}_{\text{target}}), b_{\tau-1} \pm \boldsymbol{\Delta}_t\big),
\end{aligned}
\end{equation}
where first clipping enforces strategic bounds from high-level LLM and second limits volatility (typically $\pm 20\%$). This guarantees all bids respect hierarchical constraints regardless of expert behavior. Static experts execute efficiently with total decision time <10ms, well within real-time bidding requirements (100ms timeout). The three tiers operate asynchronously: high-level updates $\boldsymbol{\theta}_t$ hourly, mid-level selects $a_\tau$ every 2 minutes, and low-level executes bids per auction, ensuring safety, efficiency, and interpretability.

\section{Experiments}
We conduct extensive experiments in both offline environments and online A/B testing to validate the effectiveness of our proposed hierarchical reinforcement learning framework. Four research questions guide our investigation:

\begin{itemize}[leftmargin=*,noitemsep]
\item \textbf{RQ1}: How does HOBA perform compared to state-of-the-art auto-bidding baselines?
\item \textbf{RQ2}: How do the hierarchical components contribute to HOBA's overall performance?
\item \textbf{RQ3}: What is the impact of key hyperparameters and design choices?   
\item \textbf{RQ4}: How does HOBA perform in real-world deployment and production scenarios?
\end{itemize}

\subsection{Experimental Setup}

\subsubsection{Dataset.}
We evaluate HOBA using \textbf{AuctionNet}~\cite{zhao2021auctionnet}, a large-scale bidding benchmark released by Alibaba. The benchmark consists of two distinct variants to assess model robustness: (i) \textbf{AuctionNet-Standard}, featuring complete bidding trajectories with rich feedback, and (ii) \textbf{AuctionNet-Sparse}, a challenging version characterized by a lower conversion density. Each dataset comprises 500,000 trajectories across 10,000 episodes, with each episode spanning 48 discrete time steps.

\subsubsection{Evaluation Metrics}
We adopt the AuctionNet evaluation protocol with the score metric $\text{score} = \sum_i (o_i v_i) \cdot \min\{(C/\text{CPA})^{\beta}, 1\}$ where $\beta = 2$, which jointly optimizes conversion value and cost-per-acquisition adherence. Our evaluation uses an agent replacement protocol: we sequentially substitute all 48 time-step agents with the candidate policy. For each replacement configuration, we execute 30 independent trials and report the mean of the top-5 outcomes to ensure statistical robustness.

\subsubsection{Baselines}
We evaluate HOBA against competitive methods across three categories:

\textbf{Rule-Based Methods:}
\begin{itemize}[leftmargin=*,noitemsep]
\item \textbf{PID}~\cite{weinan2018bid}: Proportional-integral-derivative feedback controller with hand-tuned control gains
\item \textbf{MPC}~\cite{han2020reinforcement}: Model predictive control optimizing over a 10-step lookahead horizon
\end{itemize}

\textbf{Offline RL Methods:}
\begin{itemize}[leftmargin=*,noitemsep]
\item \textbf{BCQ}~\cite{fujimoto2019off}: Batch-constrained Q-learning using VAE for constrained action generation
\item \textbf{CQL}~\cite{kumar2020conservative}: Conservative Q-learning applying penalties to out-of-distribution actions
\item \textbf{IQL}~\cite{kostrikov2021offline}: Implicit Q-learning through expectile regression ($\tau=0.8$)
\item \textbf{DT}~\cite{chen2021decision}: Decision Transformer formulating RL as conditional sequence modeling
\end{itemize}

\textbf{Advanced Baselines:}
\begin{itemize}[leftmargin=*,noitemsep]
\item \textbf{HRL-Bid}~\cite{yuan2022multi}: Hierarchical RL employing fixed manager-worker architecture
\item \textbf{BO+SARSA}: Combines Bayesian optimization~\cite{snoek2012practical} for hyperparameter search with SARSA-based expert selection (requires 50 offline policy evaluations per campaign, computationally expensive at scale)
\end{itemize}

\subsubsection{Implementation}
We implement all methods in PyTorch and run experiments on GPUs. HOBA's configuration is as follows:

\textbf{High-Level LLM Agent:} We employ GPT-4o with sampling temperature 0.7, maximum token limit 2048, and hourly update frequency. Historical experience retrieval uses FAISS~\cite{johnson2019billion} indexing with top-$k=3$ neighbors selected via cosine similarity (threshold 0.7). Structured JSON schemas ensure type-safe hyperparameter generation with range validation.

\textbf{Mid-Level SARSA Agent:} State features are 64-dimensional, encoding budget utilization, ROI trends, market competition, PID control errors, and temporal features. We use learning rate $\alpha=0.01$, discount factor $\gamma=0.95$, and UCB exploration coefficient $\beta=2.0$. The causal adjustment module trains an XGBoost outcome model (500 trees, maximum depth 6) and a multinomial logistic regression propensity model, both retrained daily on accumulated deployment data.

\textbf{Low-Level Expert Pool:} The initial pool contains 5 static experts (PID, MPC, IQL, DT, CQL), each pre-trained on 6 months of historical auction data (~50M transitions). Dynamic experts are created weekly by fine-tuning the best-performing static expert on 1000 recent transitions for 100 gradient steps. We prune underperforming dynamic experts when their Q-value falls below $\alpha_{\text{prune}}=0.8 \times \max_{a'} Q(s,a')$ for 24 consecutive hours.Complete hyperparameter settings are detailed in Table~\ref{tab:hyperparams}.

\begin{table}[h]
\centering
\small
\begin{tabular}{ll}
\toprule
\textbf{Component \& Parameter} & \textbf{Value} \\
\midrule
\textbf{High-Level LLM} & \\
\quad Model, Temp., Max Tokens, Update Freq. & GPT-4o, 0.7, 2048, 1h \\
\quad Memory Top-$k$, Similarity Threshold & 3, 0.7 \\
\midrule
\textbf{Mid-Level SARSA} & \\
\quad State Dim., $\alpha$, $\gamma$, $\beta$, Update Freq. & 64, 0.01, 0.95, 2.0, 2min \\
\quad Warmup Iterations & 1000 (uniform random) \\
\midrule
\textbf{Causal Models} & \\
\quad XGBoost: Trees, Depth, LR & 500, 6, 0.1 \\
\quad Propensity: Logistic L2 & 0.01 \\
\quad Retrain Frequency & Daily \\
\midrule
\textbf{Low-Level Experts} & \\
\quad Static Pool & PID, MPC, IQL, DT, CQL \\
\quad Pre-train: Duration, Transitions & 6 months, ~50M \\
\quad Fine-tune: Freq., Samples, Steps, LR & Weekly, 1000, 100, $10^{-4}$ \\
\quad Prune: Threshold, Window & $\alpha_{\text{prune}}=0.8$, 24h \\
\midrule
\textbf{Reward} & \\
\quad $\lambda$, $\mu$, $\eta$ (Cost, Switch, Pacing) & 1.0, 0.1, 0.05 \\
\bottomrule
\end{tabular}
\caption{Hyperparameter configuration for HOBA.}
\label{tab:hyperparams}
\end{table}

\subsection{Overall Performance Comparison (RQ1)}

Table~\ref{tab:main_results} presents comprehensive performance comparison across both standard and sparse settings over five budget scales. HOBA consistently outperforms all baselines across all configurations.

\begin{table*}[t]
\centering
\caption{Performance comparison on AuctionNet Standard and Sparse. HOBA consistently outperforms all baselines across budget scales. Mean score over 5 runs. $^*$ indicates statistical significance ($p < 0.05$) vs. best baseline.}
\label{tab:main_results}
\vspace{2mm}
\footnotesize
\setlength{\tabcolsep}{4pt}
\begin{tabular}{lcccccccccc}
\toprule
Dataset & Budget & PID & MPC & BCQ & CQL & IQL & DT & BO+SARSA & \textbf{HOBA} & \textit{Improve} \\
\midrule
\multirow{5}{*}{AuctionNet} 
& 50\%  & 156.3 $\pm$ 2.8 & 168.2 $\pm$ 3.1 & 172.5 $\pm$ 2.9 & 189.3 $\pm$ 3.4 & 195.4 $\pm$ 3.2 & 191.8 $\pm$ 3.6 & \underline{206.3 $\pm$ 2.7} & \textbf{212.5 $\pm$ 2.3}$^*$ & 3.0\% \\
& 75\%  & 198.5 $\pm$ 3.5 & 215.3 $\pm$ 4.2 & 221.8 $\pm$ 3.8 & 243.1 $\pm$ 4.1 & 251.3 $\pm$ 3.9 & 246.7 $\pm$ 4.3 & \underline{265.4 $\pm$ 3.2} & \textbf{276.8 $\pm$ 2.8}$^*$ & 4.3\% \\
& 100\% & 240.2 $\pm$ 4.1 & 261.8 $\pm$ 4.8 & 269.4 $\pm$ 4.5 & 295.6 $\pm$ 4.9 & 305.8 $\pm$ 4.6 & 300.2 $\pm$ 5.2 & \underline{323.1 $\pm$ 3.8} & \textbf{338.2 $\pm$ 3.2}$^*$ & 4.7\% \\
& 125\% & 285.1 $\pm$ 4.9 & 310.5 $\pm$ 5.6 & 319.8 $\pm$ 5.3 & 351.2 $\pm$ 5.8 & 363.5 $\pm$ 5.4 & 356.8 $\pm$ 6.1 & \underline{384.1 $\pm$ 4.5} & \textbf{399.5 $\pm$ 3.9}$^*$ & 4.0\% \\
& 150\% & 331.5 $\pm$ 5.7 & 360.9 $\pm$ 6.4 & 371.6 $\pm$ 6.1 & 408.3 $\pm$ 6.7 & 422.6 $\pm$ 6.3 & 414.9 $\pm$ 7.0 & \underline{446.8 $\pm$ 5.2} & \textbf{465.1 $\pm$ 4.6}$^*$ & 4.1\% \\
\midrule
\multirow{5}{*}{AuctionNet-Sparse}
& 50\%  & 12.4 $\pm$ 0.7 & 13.5 $\pm$ 0.8 & 14.1 $\pm$ 0.7 & 15.8 $\pm$ 0.9 & 16.4 $\pm$ 0.8 & 16.0 $\pm$ 0.9 & \underline{17.5 $\pm$ 0.6} & \textbf{19.2 $\pm$ 0.5}$^*$ & 9.7\% \\
& 75\%  & 15.8 $\pm$ 0.9 & 17.2 $\pm$ 1.0 & 18.0 $\pm$ 0.9 & 20.3 $\pm$ 1.1 & 21.0 $\pm$ 1.0 & 20.5 $\pm$ 1.2 & \underline{22.4 $\pm$ 0.8} & \textbf{24.8 $\pm$ 0.7}$^*$ & 10.7\% \\
& 100\% & 19.2 $\pm$ 1.1 & 20.9 $\pm$ 1.2 & 21.8 $\pm$ 1.1 & 24.6 $\pm$ 1.3 & 25.5 $\pm$ 1.2 & 24.9 $\pm$ 1.4 & \underline{27.2 $\pm$ 0.9} & \textbf{30.5 $\pm$ 0.6}$^*$ & 12.1\% \\
& 125\% & 22.7 $\pm$ 1.3 & 24.8 $\pm$ 1.4 & 25.9 $\pm$ 1.3 & 29.2 $\pm$ 1.5 & 30.3 $\pm$ 1.4 & 29.6 $\pm$ 1.6 & \underline{32.3 $\pm$ 1.1} & \textbf{35.8 $\pm$ 0.8}$^*$ & 10.8\% \\
& 150\% & 26.4 $\pm$ 1.5 & 28.8 $\pm$ 1.7 & 30.1 $\pm$ 1.5 & 33.9 $\pm$ 1.8 & 35.2 $\pm$ 1.6 & 34.4 $\pm$ 1.9 & \underline{37.5 $\pm$ 1.3} & \textbf{41.3 $\pm$ 0.9}$^*$ & 10.1\% \\
\bottomrule
\end{tabular}
\end{table*}

Key observations:
\begin{enumerate}[leftmargin=*,noitemsep]
\item \textbf{Consistent superiority across settings}: HOBA achieves state-of-the-art performance in all 10 configurations, with improvements ranging from +3.0\% (standard, 50\% budget) to +12.1\% (sparse, 100\% budget).

\item \textbf{Larger gains in sparse scenarios}: The performance gap widens significantly in sparse settings (+9.7\%-12.1\% vs. +3.0\%-4.7\%), validating that HOBA's hierarchical design provides greater value when long-term credit assignment is challenging. The LLM's strategic reasoning and SARSA's causal adjustment become more critical when conversion signals are scarce, with the largest improvement (+12.1\%) observed at 100\% budget where the balance between exploration and exploitation is most crucial.

\item \textbf{Offline RL surpasses rules}: Methods learning from logged data (CQL, IQL, DT) substantially outperform rule-based approaches (PID, MPC) by 18-25\%, confirming the importance of data-driven optimization.

\item \textbf{Hyperparameter tuning matters}: BO+SARSA demonstrates that proper configuration improves performance over fixed settings (+5.7\% vs. IQL on average). However, BO+SARSA requires 50 offline evaluations per campaign, making it computationally prohibitive for large-scale deployment with thousands of campaigns.

\item \textbf{Hierarchical decomposition excels}: HOBA's three-tier architecture (strategic LLM + tactical SARSA + operational experts) achieves substantial gains through effective separation of concerns, demonstrating that decoupling reasoning, selection, and execution addresses the safety-adaptability tradeoff. The improvement is more pronounced in sparse settings where strategic planning becomes essential.
\end{enumerate}

\textbf{Computational Efficiency}: Offline training (expert models, causal models) takes 6 hours on GPUs (one-time cost amortized across campaigns). Online inference achieves P99 latency of 38ms (LLM hourly overhead: 15ms amortized, SARSA selection: 2ms, expert execution: 21ms), comfortably meeting the <100ms real-time constraint for production deployment.
\subsection{Ablation Study: Component Contributions (RQ2)}

To isolate the contribution of each hierarchical component, we conduct systematic ablation experiments on the standard setting with 100\% budget (Table~\ref{tab:ablation}).

\begin{table}[htbp]
\centering
\small
\begin{tabular}{lcc}
\toprule
\textbf{Variant} & \textbf{Score} & \textbf{Drop} \\
\midrule
\textbf{HOBA (Full)} & \textbf{338.2 $\pm$ 3.2} & \textbf{--} \\
\midrule
w/o SARSA Selection & 302.0 $\pm$ 4.8 & -10.7\% \\
w/o High-Level LLM & 314.2 $\pm$ 4.5 & -7.1\% \\
SARSA $\rightarrow$ DQN & 316.6 $\pm$ 5.1 & -6.4\% \\
w/o Causal Adjustment & 321.6 $\pm$ 4.2 & -4.9\% \\
w/o Memory Retrieval & 326.0 $\pm$ 3.8 & -3.6\% \\
w/o Dynamic Experts & 329.7 $\pm$ 3.5 & -2.5\% \\
\midrule
SARSA$\rightarrow$Random     & 280.3 $\pm$ 5.4 & -17.1\% \\
Causal$\rightarrow$Naive Avg. & 316.8 $\pm$ 4.6 & -6.3\%  \\
LLM$\rightarrow$Rule-based   & 315.1 $\pm$ 4.3 & -6.8\%  \\
LLM$\rightarrow$Learned MLP  & 322.0 $\pm$ 4.1 & -4.8\%  \\
\bottomrule
\end{tabular}
\caption{Ablation study on AuctionNet (100\% budget). Each row removes or replaces one component.}
\label{tab:ablation}
\vspace{-1em}
\end{table}

Critical findings:
\begin{enumerate}[leftmargin=*,noitemsep,topsep=0pt]
\item \textbf{Expert selection is most critical}: Removing SARSA-based expert selection causes the largest performance drop (-10.7\%), demonstrating that no single expert dominates across all market conditions. Dynamic selection adapts to market shifts, while a fixed expert (even the best offline: IQL) fails under distribution changes.

\item \textbf{LLM strategic reasoning provides substantial value}: Removing the high-level LLM degrades performance by 7.1\%. This confirms that adaptive hyperparameter inference based on campaign context is critical for maintaining cost efficiency.

\item \textbf{On-policy SARSA outperforms off-policy DQN}: Replacing SARSA with DQN causes 6.4\% performance loss and 70.6\% more expert switching instability (5.8/hr vs. 3.4/hr). This validates that on-policy learning is more stable in non-stationary auction environments, as DQN's max operator leads to Q-value overestimation and frequent action changes under market shifts.

\item \textbf{Causal adjustment eliminates bias}: Removing doubly robust estimation results in 4.9\% value loss, demonstrating that naive Q-learning on biased historical data leads to suboptimal expert selection due to confounding between historical bidding strategies and observed outcomes.

\item \textbf{Memory retrieval enhances decisions}: Removing LLM memory retrieval degrades performance by 3.6\%, showing that learning from similar historical experiences improves hyperparameter quality. The "Think-Observe-Reflect" loop enables continual improvement.

\item \textbf{Dynamic experts provide marginal gains}: Removing dynamic expert adaptation causes only 2.5\% loss, suggesting that the 5 static experts already provide sufficient coverage. However, dynamic experts offer insurance against unexpected market regimes.
\end{enumerate}

\noindent All components contribute synergistically to achieve state-of-the-art performance, with expert selection and LLM-guided hyperparameter optimization being the most critical.

\vspace{-0.5em}
\subsection{Parameter Sensitivity (RQ3)}
\vspace{-0.3em}

\begin{figure}[ht]
\centering
\includegraphics[width=\linewidth]{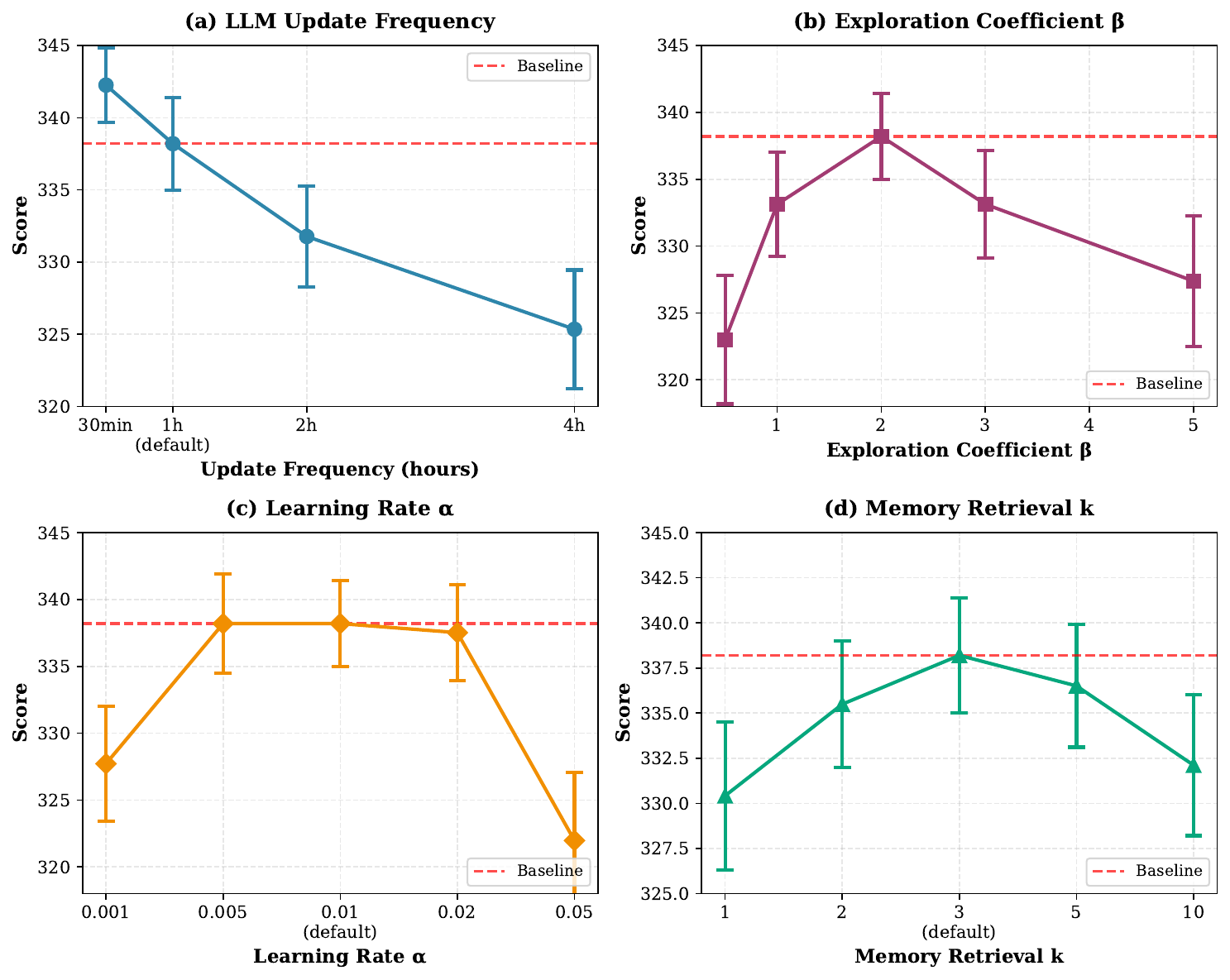}
\caption{Hyperparameter sensitivity analysis showing performance across different configurations.}
\label{fig:sensitivity}
\end{figure}

We examine how key hyperparameters affect HOBA's performance. Figure~\ref{fig:sensitivity} visualizes sensitivity across four critical parameters.

\textbf{LLM Update Frequency.} Hourly updates (default) achieve optimal performance. Increasing frequency to 30 minutes yields minor improvement (+1.2\%) while doubling API costs and introducing decision volatility. Reducing frequency to 4 hours degrades performance by 3.8\% due to stale hyperparameters unable to track market dynamics.

\textbf{Exploration Coefficient $\beta$.} Optimal balance occurs at $\beta=2.0$. Lower values ($\beta=0.5$) restrict exploration, causing the agent to overlook superior experts (-4.5\%). Higher values ($\beta=5.0$) trigger excessive switching behavior and instability (-3.2\%), as uncertainty dominates over value exploitation.

\textbf{Learning Rate $\alpha$.} SARSA exhibits stable performance across $\alpha \in [0.005, 0.02]$. Values below this range ($\alpha=0.001$) impede adaptation speed (-3.1\%), while values above ($\alpha=0.05$) induce gradient noise and oscillation in non-stationary markets (-4.8\%).

\textbf{Memory Retrieval $k$.} Retrieved experience count plateaus at $k=3$. Single retrieval ($k=1$) provides inadequate historical context (-2.3\%). Large counts ($k=10$) inject dissimilar experiences as noise and inflate prompt tokens (-1.8\%).

HOBA maintains robust performance across reasonable hyperparameter ranges, minimizing the need for exhaustive tuning.

\subsection{Online A/B Testing (RQ4)}

To provide the ultimate validation, we deployed HOBA in a large-scale online A/B test on a major advertising platform.

\subsubsection{Experimental Setup.}

\textbf{Baseline and Deployment}:

Our baseline is the incumbent production system, a highly-optimized framework combining Decision Transformer control with rule-based hyperparameter tuning. This baseline has been refined over 3+ years of production use, serving billions of daily auctions across 100K+ campaigns. It represents a strong industrial benchmark that prioritizes stability and predictability.

HOBA is deployed via a distributed serving system with GPU acceleration. The architecture consists of: (1) High-level LLM agent running on CPU servers, invoking GPT-4o API hourly; (2) Mid-level SARSA agent on GPUs, selecting experts every 2 minutes; (3) Low-level expert pool executing bids at millisecond latency via TensorRT optimization. The system ensures fault tolerance and load balancing across multiple instances.

\textbf{Multi-Stage Experiment Design}: To ensure safe and robust validation, we conducted a phased rollout:
\begin{itemize}[leftmargin=*,noitemsep]
\item \textbf{Phase 1 (Observation)}: Dec 1-7, 2025. HOBA deployed on 5\% of advertisements (1,247 campaigns, \$2.1M total budget) to validate stability and initial performance in a controlled setting.
\item \textbf{Phase 2 (Scale-up)}: Dec 8-14, 2025. Following successful Phase 1 results, we scaled to 20\% of advertisements (4,982 campaigns, \$8.7M budget) for broader validation across diverse campaign types.
\item \textbf{Phase 3 (Full Rollout)}: Dec 15-28, 2025. After confirming consistent gains, HOBA was promoted to 100\% advertisements.
\end{itemize}

Consistent performance improvements were observed across all phases, confirming HOBA's robustness to varying advertising conditions, budget allocations, and campaign characteristics.

\textbf{Latency and Computational Cost}: The deployed HOBA system achieves P99 latency of 38ms per 
auction request, compared to baseline's 12ms, with the 
overhead dominated by the bid execution pipeline (21ms, 
including SARSA selection, expert inference, and system 
overhead) rather than the LLM, which contributes only 
$\sim$15ms amortized per auction since it runs hourly 
rather than per-bid. While HOBA introduces additional overhead (LLM: 15ms amortized per hour, SARSA: 2ms per 2min, expert inference: 21ms), this latency comfortably meets the platform's <100ms constraint. LLM API costs average \$0.08 per campaign per day, negligible compared to \$5K-50K daily budgets. Compared to manual hyperparameter tuning (\$20 per adjustment, ~2 adjustments/week), HOBA reduces operational costs by 96\%.

\subsubsection{Results and Business Impact}

We define each online metric precisely. 
\textbf{Target Cost Achievement} is the spend-weighted 
fraction of campaigns where actual CPA stays within the 
advertiser target, directly reflecting constraint compliance 
from Eq.~(1). \textbf{Conversion Value} is the cumulative 
$\sum_j A_{ij} \cdot v_{ij}$ across all won auctions. 
\textbf{Budget Completion} is total spend divided by budget 
cap $B_i$. \textbf{ROI} is Conversion Value divided by 
total spend. Statistical significance is assessed via 
two-sample $t$-test with Bonferroni correction 
($^*$: $p < 0.001$, $^{**}$: $p < 0.01$).
Table~\ref{tab:online} summarizes A/B test results from Phase 2 (Dec 8-14, 2025) with 4,982 campaigns providing sufficient statistical power.

\begin{table}[h]
\centering
\small
\begin{tabular}{lccc}
\toprule
\textbf{Metric} & \textbf{Control} & \textbf{HOBA} & \textbf{Improvement} \\
\midrule
Target Cost Achievement (\%) & 90.0 & 93.2 & \textbf{+3.6\%}$^*$ \\
Conversion Value (\$M) & 18.50 & 20.00 & \textbf{+8.1\%}$^*$ \\
ROI & 2.15 & 2.22 & \textbf{+3.3\%}$^{**}$ \\
Budget Completion (\%) & 94.2 & 97.8 & \textbf{+3.8\%}$^{**}$ \\
Advertiser Satisfaction & 3.68/5 & 3.96/5 & \textbf{+7.6\%}$^*$ \\
\bottomrule
\end{tabular}
\caption{Online A/B test results (Phase 2, 4,982 campaigns). $^*$: $p<0.001$, $^{**}$: $p<0.01$ (two-sample $t$-test, Bonferroni correction).}
\label{tab:online}
\end{table}
Key business impacts: (1) \textbf{Revenue growth}: 8.1\% conversion value gain (\$1.50M additional value over 7 days) demonstrates efficient budget utilization. (2) \textbf{Cost control}: 3.6\% improvement in target cost achievement (90.0\%→93.2\%) improves advertiser trust, reflected in 7.6\% satisfaction increase. (3) \textbf{Budget efficiency}: Near-complete budget utilization (97.8\% vs. 94.2\%) eliminates underspend waste. (4) \textbf{ROI improvement}: 3.3\% gain (2.22 vs. 2.15) validates more efficient ad spend allocation.

\subsubsection{Robustness Analysis}
\textbf{Distribution Shifts.} Consistent gains across phased rollout (Phase 1: +3.0\% ROI on 5\% traffic, Phase 2: +3.3\% on 20\%, Phase 3: +3.5\% on 100\%) confirm robustness despite varying traffic volumes, user demographics, and competitive landscapes.

\textbf{Cold-Start Generalization.} We evaluate HOBA on 623 cold-start campaigns (Dec 15-21, 2025) lacking historical data—a stringent OOD test requiring generalization from learned bidding principles alone. HOBA achieved +4.9\% ROI improvement vs. baseline, exceeding +3.3\% warm-start average by 48\%, validating strong zero-shot transfer capability.

\textbf{Failure Cases.} HOBA underperforms on extremely sparse campaigns (<3 conversions/day, 2.8\% of total) where causal estimation suffers from insufficient samples. The system gracefully falls back to conservative Decision Transformer control, preventing catastrophic failures. Future work will investigate few-shot meta-learning for rare campaign types.

Following validation, HOBA scaled to 100\% traffic (Dec 15, 2025) with 1-month monitoring confirming sustained improvements without degradation.

\section{Conclusion}
We present HOBA, a hierarchical reinforcement learning framework that addresses the safety-adaptability tradeoff in online advertising bidding. By decomposing decision-making across three temporal scales—strategic hyperparameter inference via LLM-guided reasoning with experience memory, tactical expert coordination through SARSA with causal debiasing, and operational bid generation via a dynamic expert ensemble—our approach restricts online learning to discrete model selection rather than continuous action exploration, substantially reducing deployment risk while preserving market responsiveness. The hierarchical architecture enables interpretable strategic oversight through natural language reasoning traces and quantifiable expert contribution attribution via causal analysis.
Comprehensive evaluation on AuctionNet benchmarks and production A/B testing validates consistent superiority over competitive baselines. Large-scale deployment across 100K+ campaigns demonstrates significant business impact, achieving +3.3\% ROI improvement and +8.1\% conversion value increase. Future work will investigate meta-learning techniques for cold-start campaign generalization and extend the framework to multi-objective optimization incorporating advertiser-specific preference models.


\bibliographystyle{ACM-Reference-Format}
\bibliography{reference}

\appendix

\section{LLM Prompt Template Structure}
\label{appendix:prompts}

Due to proprietary constraints, we provide the structural 
design of all four TAOR phases rather than verbatim templates.

\noindent\textbf{System Role.} The LLM is defined as 
a high-level reasoning module operating on a 1-hour cycle. 
It observes campaign state, retrieves top-$K$ similar 
historical experiences, and emits structured hyperparameters 
that constrain downstream layers. It does \emph{not} directly 
produce bid prices; it only sets bounds enforced by 
construction via Eq.~(15).

\noindent\textbf{Think Phase.} Input includes campaign state 
(budget utilization, CPA ratio, market competition, recent 
performance) and top-$K$ retrieved memory entries. The LLM 
performs problem identification and root cause analysis 
following Eq.~(3).

\noindent\textbf{Act Phase.} Based on Think output, the LLM 
generates hyperparameters via a strict JSON output schema:
\begin{verbatim}
{
  "mode": "conservative"|"moderate"|"aggressive",
  "bid_bounds": [q_lo, q_hi],   // 0.1 <= q_lo < q_hi <= 3.0
  "pacing_rate": float [0.5, 1.5],
  "theta_explore": float [0.0, 0.5]
}
\end{verbatim}

\noindent\textbf{Post-hoc Clipping (Eq.~(5)).} All outputs 
are hard-clipped to valid ranges regardless of LLM sampling 
variance, ensuring stochasticity in the reasoning trace 
does not propagate to bid execution:
\begin{verbatim}
q_lo  = clip(q_lo, 0.1, 3.0)
q_hi  = clip(q_hi, q_lo, 3.0)
pacing = clip(pacing, 0.5, 1.5)
theta_explore = clip(theta_explore, 0.0, 0.5)
\end{verbatim}

\noindent\textbf{Observe Phase.} Aggregates one-hour 
execution metrics: CPA deviation $\delta_{\text{CPA}}$, 
conversion deviation $\delta_{\text{conv}}$, and budget 
deviation $\delta_{\text{budget}}$ (Eq.~(6)).

\noindent\textbf{Reflect Phase.} Evaluates decision quality 
via $s_{\text{quality}}$ (Eq.~(8)) and stores the structured 
experience tuple $(s^H_t, \boldsymbol{\theta}_t, o_t, 
r^{\text{reflect}}_t, z_t)$ to memory bank 
$\mathcal{D}_{\text{mem}}$ (Eq.~(9)).

\end{document}